\documentclass[sigconf,nonacm,table]{acmart}

\usepackage{xcolor}

\usepackage{diagbox}

\AtBeginDocument{%
  \providecommand\BibTeX{{%
    \normalfont B\kern-0.5em{\scshape i\kern-0.25em b}\kern-0.8em\TeX}}}

\setcopyright{acmcopyright}
\copyrightyear{2020}
\acmYear{2020}
\acmDOI{10.1145/1122445.1122456}

\acmConference[ACM SIGSPATIAL 2020]{ACM SIGSPATIAL 2020: International Conference on Advances in Geographic Information Systems 2020}{November 03--06, 2020}{Seattle, WA}
\acmBooktitle{ACM SIGSPATIAL 2020: International Conference on Advances in Geographic Information Systems 2020, November 03--06, 2020, Seattle, WA}

\begin{document}

\title{Synthetic Data and Hierarchical Object Detection in Overhead~Imagery}

\author{Nathan~Clement, Alan~Schoen, Arnold~Boedihardjo, and Andrew~Jenkins}
\authornote{All authors are affiliated with Maxar Technologies Inc. (formerly DigitalGlobe), Herndon, Virginia.}
\email{{nathan.clement, alan.schoen, arnold.boedihardjo, andrew.jenkins}@maxar.com}

\renewcommand{\shortauthors}{Clement, et al.}

\begin{abstract}
The performance of neural network models is often limited by the availability of big data sets. To treat this problem, we survey and develop novel synthetic data generation and augmentation techniques for enhancing low/zero-sample learning in satellite imagery. In addition to extending synthetic data generation approaches, we propose a hierarchical detection approach to improve the utility of synthetic training samples. We consider existing techniques for producing synthetic imagery--3D models and neural style transfer--as well as introducing our own adversarially trained reskinning network, the GAN-Reskinner, to blend 3D models. Additionally, we test the value of synthetic data in a two-stage, hierarchical detection/classification model of our own construction. To test the effectiveness of synthetic imagery, we employ it in the training of detection models and our two stage model, and evaluate the resulting models on real satellite images. All modalities of synthetic data are tested extensively on practical, geospatial analysis problems. Our experiments show that synthetic data developed using our approach can often enhance detection performance, particularly when combined with some real training images. When the only source of data is synthetic, our GAN-Reskinner often boosts performance over conventionally rendered 3D models and in all cases the hierarchical model outperforms the baseline end-to-end detection architecture.
\end{abstract}


\maketitle

\section{Introduction}

Applications of object localization and classification often involve working with limited training data. Synthetic data offers a possible way to train models when the available number of training samples is very small. Even when thousands of training samples are available, synthetics can improve model accuracy. Additionally, manually labeling training data is tedious, time consuming, and cost prohibitive, so reducing the number of samples required for generating a well-performing model is desirable.

To understand the value of synthetic training data for our domain, object detection in overhead imagery, we analyzed and conducted extensive tests of various forms of manufactured data. 3D Computer-aided Design (CAD) models have been a popular source of synthetic data for training computer vision models. We test this conventional technique as well as our proposed generative adversarial network (GAN) based technique, GAN-Reskinner, for improving 3D synthetic data, and a neural style transfer based technique for data augmentation.

An important advantage of 3D CAD models is the high level of flexibility it affords for synthesizing a training image. With 3D CAD models, an arbitrary 3D object can be constructed and rendered with light source and camera at any given angles. Prior works have shown that images rendered from 3D CAD models can make good training data for object detection networks: \cite{orig3d,nvidia3d1,finetune3d,nvidia3d2}. In practice one cannot always match the performance of real training data by using synthetic data alone; this gap in training utility is sometimes refereed to as `domain shift' or the `reality gap' \cite{domainShiftSeg}. One possible explanation for this gap is that in a synthetic image, the 3D models are too easily distinguishable from the background. To remedy this problem, we devised the GAN-Reskinner to take a conventional 3D CAD image and re-render it to be more fully integrated into the scene using GAN based in-painting. We find that this approach does improve the training utility of the synthetic data in certain cases.

Another problem of working with sparse training data is correlation with irrelevant imaging conditions. For example: if an object detection model is trained on images collected over a city in July, it may do well at generalizing to a new city image captured in the summer, but fail on images over a snowy countryside area. In other words, machine learning models trained with low sample size are more susceptible to overfitting to a subdomain of satellite imagery. To alleviate this problem, we use neural style transfer \cite{nnstx} to augment small datasets in such a way that the resulting model is more robust to different imaging conditions. 

Each of the above techniques addresses different technical aspects related to generating effective synthetic data. Our approach is to integrate all of the above - conventional 3D object generator, GAN-Reskinner, and neural style transfer - into a unified system that is more flexible and produces significantly higher quality training samples than any single data manufacturing approach. Additionally, we propose an object detection architecture that exploits our synthetic data to improve performance on low sample regimes. We devise a framework that employs a multi-stage filtering approach: a broad class detector stage from which outputs are filtered by a specific/narrow class filtering stage. We call this model \textit{broad-to-narrow}. We find that this multi-stage approach greatly improves detection quality especially in the case of zero-shot learning, where real training data from closely related classes can help the \textit{broad} detection stage, and the synthetic data of the target class can be used to train the second \textit{narrow} stage model. For example, if our target class is compact passenger cars then the broad stage will detect and output small/medium/large passenger cars, buses, trucks, etc., and the narrow stage model will filter those outputs for compact passenger car target class.

We summarize our key contributions below:

\begin{itemize}
\item GAN-Reskinner, a novel synthetic data generation technique for overhead imagery that exploits the flexibility of 3D CAD models and robust texture generation of adversarially trained networks. The resulting images blend synthetic objects in a more integrated fashion that can better inform discriminative features to the detection model.
\item Unified synthetic data generation approach that utilizes GAN-Reskinner, 3D CAD models, and neural style transfer to increase the diversity of complementary features in synthetic scenes and improve training data utility.
\item The \textit{broad-to-narrow} detection architecture that utilizes a two-stage object recognition model to leverage synthetic data and enhance detection performance over existing state of the art end-to-end\footnote{We use the term end-to-end to specify a single detection model, like Faster R-CNN, fine tuned in one training run on the detection dataset.} approaches for low sample and zero-shot settings. We also provide extensive analysis of the \textit{broad-to-narrow} detection model to show its improved detection quality over state of the art approaches. 
\item Comprehensive experiments to test and evaluate the effectiveness of the unified synthetic data generation approach and \textit{broad-to-narrow} model.  
\end{itemize}

The remainder of this paper is organized as follows. Section \ref{related_work} describes the related works and highlights important technical gaps of existing methods. Section \ref{methods} presents the technical details of our proposed approach for synthetic image generation and the \textit{broad-to-narrow} detection model. Section \ref{results} provides the results of our extensive experiments that test the effectiveness of various synthetic generation strategies and \textit{broad-to-narrow }model applied to low sample and zero-shot learning settings and presents analysis of our \textit{broad-to-narrow} detection model. Lastly, Section \ref{conclusion} provides our conclusion and future work. 

\section{Related Work}
\label{related_work}

\subsection{Synthetic training images and domain randomization}

The use of rendered 3D models as training data has proved valuable for training object detection models \cite{orig3d,nvidia3d1,finetune3d,nvidia3d2}. In all these works, some notion of `domain randomization' has been applied to object detection. The essential idea of domain randomization is that the difference between real imagery and 3D generated scenes may be overcome by sufficient randomization of the scene elements. Paint patterns and colors and image backgrounds are all permuted in the interest of domain randomization. As a result, the only attribute held constant over a training set of synthetic data is the essential underlying geometry of the objects of interest. This observation informs our strategy for modelling synthetic data to emphasize discriminative features.

A much different approach to data simulation is taken by the Digital Imaging and Remote Sensing Image Generation (DIRSIG) \cite{dirsig} system. Their rendering model works to create realistic overhead imagery via complex physical simulations. The DIRSIG system has been used to generate imagery to train a helicopter detector, and thus increased model recall from 30\% to 60\% \cite{dirsigTest}.

\subsection{Domain adaptation}

Much work has been done to address the domain adaptation problem and specifically targeting the gap between synthetic and real data. A typical component of approaches to domain adaptation is the matching of the sets of deep features extracted from source and target domains. 
For example, in \cite{ganin2014unsupervised} the authors include an adversarial network to learn and minimize the discrepancy between source and target features.
Closer to our task of object detection, the authors of \cite{domainShiftSeg} attacked the problem of domain adaptation for pixel segmentation in natural imagery by a related adversarial approach.

A different approach is taken by SimGAN \cite{simgan} to close the gap between real and synthetic image distributions. They train a discriminator which learns to quantify the domain gap, and a generator which learns to refine synthetic images to close that domain gap \textit{at the image level}. They demonstrate the effectiveness of this refinement process for improving synthetic training data in gaze and pose estimation problems.

The authors of \cite{simGanSat} employ the SimGAN technique to improve the realism of 3D CAD models of aircraft in overhead imagery. Though potentially useful for training a classifier, the image chips produced are too small and regular for effective detector training. The authors mark the improvement of their refined synthetic data by a decreased maximum mean discrepancy to real data, but not by its utility as training data. Our proposed GAN-Reskinner can produce outputs of any given image size and fidelity to serve as effective detector training data. 

Since we are interested in developing techniques for low sample learning, the domain adaptation problem is salient. Our GAN-Reskinner can be viewed as a variant of domain adaptation in which the semantic representation (meant to be divorced from any particular domain) is given explicitly.

\subsection{Neural style transfer}
The concept of neural style transfer was introduced by Gatys et al. and uses Gram matrices of hidden layer feature representations from the VGG network to quantify a notion of style \cite{nnstx}. Their neural algorithm of artistic style (NAAS) provides a way to capture both the semantic content of an image and the image's style and furthermore to mix the style and content of two different images. In the paper, the authors demonstrate the algorithm's impressive ability to impose the style of a particular painting on an arbitrary scene. For example, a photograph of a canal is rendered in the style of `The Starry Night' by Vincent van Gogh. 

The NAAS works by using a convolutional neural network to specify a style distance and content distance on image space. 
The content distance measures how close two images are to representing the same semantic content, while the style distance measures how close two images are in style.
After choosing content and style images $A$ and $B$, the array that will become the output, $X$, is initialized $X = A$. Training consists of optimizing the pixels of $X$ to simultaneously minimize $d_{\mbox{\tiny content}}(X,A)$ and $d_{\mbox{\tiny style}}(X,B)$. After convergence, $X$ is deemed the transfer of $B$'s style onto $A$'s content.

Our interest is not in using neural style transfer to capture the style of various artists, but rather to capture the (otherwise difficult to quantify) qualitative variations present in satellite imagery. We use the NAAS as a means of augmenting a small data set in order to achieve improved low sample learning.

\subsection{GAN-based synthetics}
With the advent of the generative adversarial network \cite{gan} came new possibilities for the manufacture of synthetic data. 
In the domain of medical imaging, \cite{frid2018synthetic} were able to implement a deep convolutional GAN (DCGAN) \cite{dcgan} which produced 64x64 pixel images to improve the performance of their classification models. Toward the task of brain scan segmentation, \cite{bowles2018gan} were able to use the technique of progressive growing of GANs \cite{pggan} to improve their segmentation models' performance, especially in the lower sample regime.

These two papers produced synthetic images of size 64x64 and 128x128, respectively. To produce larger size GAN synthetics, the approach taken by \cite{pix2pix}, `pix2pix', has been used successfully by many researchers. The pix2pix GAN is conditional and casts the GAN generator as an image-to-image translator.
For example, they translate maps into overhead images, gray scale images to color, and daytime landscape scenes to nighttime ones.

Image to image translation is a powerful idea for data generation, but in order to apply it, one is still left with the task of choosing input images. For example, to create images of shoes, the authors first had to procure edge detects of all the shoe images in their training set. From the perspective of data generation, one would then have to generate new line drawings of shoes to create new images of shoes. The authors of \cite{guibas2017synthetic} solve this problem by using two GAN stages toward the task of generating retina images together with blood vessel segmentation masks. Their pix2pix implementation is designed to take blood vessel segmentation masks and produce color retinal images. To produce new images, they use a DCGAN to generate realistic segmentation masks (without color image) then feed these into the second stage pix2pix GAN. Their two stage system was able to nearly match the training utility of real data.

A closely related work to our area of research is the pedestrian synthesis GAN of \cite{ouyang2018pedestrian}. Their system involves a GAN with two discriminators which is capable of filling in empty regions in a street scene with synthetic pedestrians. They test their system by using their GAN synthetics mixed with the ground truth training set to train a pedestrian detector. They show improvement with the inclusion of some synthetic data in the training set. A limitation of this method is that it requires real training images, whereas our GAN reskinning system works in the zero-shot setting.

Another pix2pix based system is built and used for training a car detector in \cite{howe}.
Their GAN reproduces overhead urban scenes, including their target object, from full semantic segmentation information.
Though they show good results in the low sample range, their method requires real training data of the target object. Additionally, their method incurs a high human supervision cost since it requires fully segmented scenes for training, whereas we require a system capable of quickly producing useful training data with little human supervision. In contrast, our GAN-Reskinner produces its output scenes based on an automatically computed schematic representation of satellite imagery and accepts as input arbitrary, detailed geometries of an unseen target class.

\subsection{Ensembles}

Various model ensemble schemes have shown promising results in improving the accuracy and uncertainty calibration of deep learning models. The linearly weighted super learner of \cite{ju2018relative} outperforms both majority voting and the best base learner when combining the classification predictions of convolutional neural networks (CNNs) on the CIFAR-10 dataset. Further, the authors demonstrate that even adding weak learners can improve the overall performance of an ensemble. Along with the chief goal of improving the calibration of model uncertainty, \cite{lakshminarayanan2017simple} manages to improve accuracy as well by simply averaging their base learners' probability scores. In these examples, all base learners are trained to perform the same task, but are able to attenuate each others' mistakes. This notion of ensemble seeks to make use of `diversity of errors'.

A form of ensemble (`ensemble of specialists') is described in \cite{hinton2015distilling}. The paper introduces a hierarchical classification model which ensembles a general classifier together with many specialist classifiers which may be called upon in case the general classifier indicates a test sample is part of a confusing subset of the label classes. They show an improvement in test accuracy on the large, massively multi-class Google JFT-300M data set.

Our \textit{broad-to-narrow} model follows a similar hierarchical structure, but the generalist stage is an object detection model rather than a classifier. Additionally, our approach optimizes low-shot and zero-shot settings by tailoring the synthetic data mixtures differently for the broad and narrow models and combines the confidence scores of the two-stage models to make its final confidence judgment. This ensemble strategy allows us to reap the benefits of higher recall detection models without sacrificing precision.

\subsection{Low-shot and zero-shot learning}

The problems of low-shot learning (few training examples of target class) and zero-shot learning (no training examples of target class) have spawned a very active body of research. Successful methods include transfer learning from large data sets \cite{transferable} and training models to make predictions based on a similarity metric \cite{cosine1,cosine2}. 

After training a CNN with millions of parameters on a large data set such as ImageNet, there are a few approaches to transferring the network's learned knowledge. Simplest is to use the CNN without its last (fully connected) layer to turn images into feature vectors, then to train a simple classifier (most typically a logistic regression) on those vectors. When this is done without modifying the body of the CNN, we refer to the method as \textit{feature extraction}. We reserve the name \textit{finetuning} for the related practice of allowing all layers of the network to continue to train towards minimizing the usual cross entropy loss assessed on the smaller target data set. In both cases, the final classification layer is implemented as a learnable affine linear transform. 
Some researchers have found better results by using a cosine similarity layer instead of affine linear \cite{cosine1,cosine2}. In our work, finetuning classifiers for our \textit{broad-to-narrow} architecture gave much better results than the static feature extraction method.
\enlargethispage{\baselineskip}

\section{Methods}
\label{methods}

\subsection{Overview}

\begin{figure}
    \centering
    \includegraphics[width=15em]{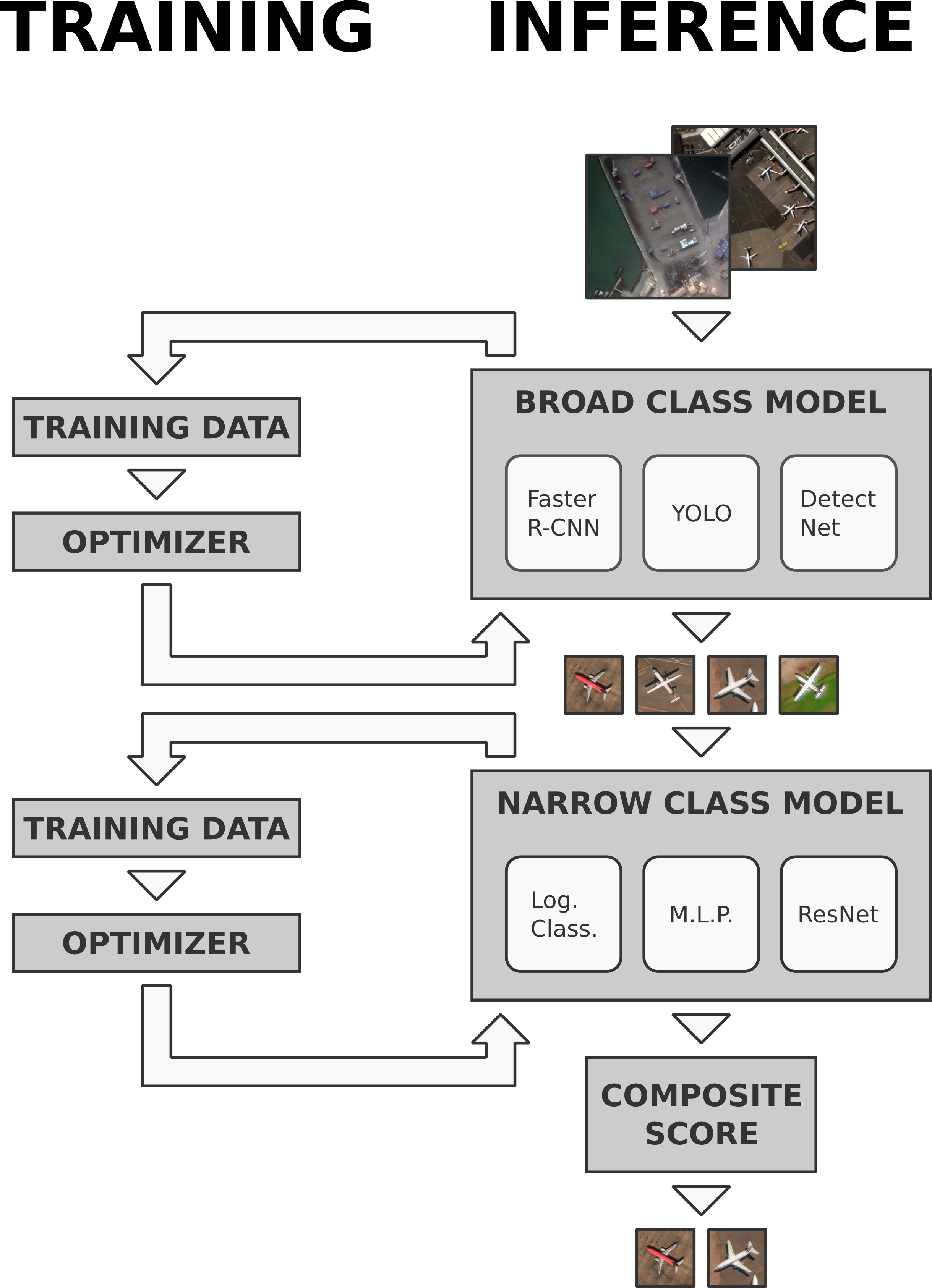}
    \caption{Diagram of our proposed two stage model.}
    \label{b2n_diagram}
\end{figure}

Our application domain is low/no sample object detection in overhead imagery. With little to no training data, we need to produce object detection models which can operate effectively over large rasters of satellite data and identify objects matching a narrow classification. Often times the objects we seek in satellite imagery are part of a hierarchical taxonomy and we can use this extra knowledge to shape our approach to the problem. For example, if we desire to detect all objects of type \texttt{tug boat} we might make use of all the data containing objects within the superclass \texttt{boat}. This is the idea driving our \textit{broad-to-narrow} approach: since it can be difficult to obtain high recall performance from a detector for a very specific class, we first design a detector to collect many candidate boxes targeting a more general class, then we filter for the sub-class of interest using a powerful secondary classifier.

Our approach is illustrated by Fig. \ref{b2n_diagram}. The training of the two stages may be done in parallel. Each model receives its own set of training images\textemdash larger crops for the detector and smaller for the classifier.\footnote{These training sets may safely be taken from the same pool of data, unlike some ensemble training methods which require multiple disjoint training sets.} The corresponding optimization problems are solved in parallel. In contrast, inference mode is sequential. First the parent class detector is run over a large swath of satellite imagery. All resulting detect boxes are next used to crop small chips from the original image strip which are passed to the classifier. The classifier gives its confidence on each of these small chips, and finally the detector and classifier confidence are combined to give a final, composite score to each detect.
See section \ref{b2n_opt} for details on the compositing of the final score.

Our process for creating training datasets 
consists of selecting samples of real, 3D, and GAN-Reskinner images and combining them at some ratio. In addition we can use neural style transfer to amplify the dataset at this point. When we train two stage models, we have not one but two datasets to build, which gives us the flexibility to chose the most suitable types of data for the localization and classification tasks independently.

When we build datasets for the zero-shot scenario, we can employ real satellite imagery containing sub-classes other than the target one. This will bias the detection stage in favor of the other sub-classes, but the bias will be rectified by the classification stage which can use synthetic data from the target sub-class to prune unwanted detects.

\subsection{Broad-to-Narrow}

A key goal of the \textit{broad-to-narrow} model is to ensure that the broad stage model outputs candidate objects with sufficiently accurate localization in order to maximize the effectiveness of the narrow filtering stage. The narrow model will have been trained to discriminate centered or nearly centered object instances, so we need the broad stage to predict accurate bounding boxes; we also want the broad stage to minimize false positive outputs in order to optimize overall precision of the final predictions.

The broad stage model is implemented using Faster R-CNN \cite{fasterrcnn} with a ResNet-50 based feature pyramid network (FPN) as the feature extractor. Using available pre-trained weights for the feature extraction and RPN stages, this state-of-the-art detector can effectively be tuned to fairly small data sets. However, if the number of training annotations is low i.e., in the hundreds or less, or if the only available data is synthetic, the model's generalization performance can degrade significantly. 
For comparison to broad-to-narrow, we also finetune the detection network end-to-end using only the target sub-class annotations.

For training and inference, large GeoTIFFs are decomposed into manageable sized `chips' with overlap. At test time, we run over large areas of interest with relatively low annotation density. Results of chip-by-chip inference are re-assembled into geospatial coordinates and non-maximum suppression is used to combine predictions on chip overlaps.

In the case of two sub-classes of a parent class, it is typical to find that two single class detectors outperforms a single two class detector. This is possibly related to class imbalance issues. In any case, when training an end-to-end detection model (as distinct from our \textit{broad-to-narrow }models) we only label objects of the target sub-class as positive in the training images.

The \textit{broad-to-narrow} strategy is flexible as it can support additional layers of classifiers at the post-detection stage, for example to further prune false positives from the detector. Another possibility is to ensemble classifiers after the detection stage. We are afforded the flexibility to employ different classification schemes by handling object localization in one robust general detection step. Since the training of the detector and classifier or classifiers in the \textit{broad-to-narrow} model are independent, we can adjust or completely replace the second stage if a better classification model is discovered. We see this modularity as a major benefit of the \textit{broad-to-narrow} approach.

\subsection{Optimizations for Broad-to-Narrow}
\label{b2n_opt}
Due to our applications of low-shot or zero-shot scenarios, the \textit{broad-to-narrow} model architecture was designed to produce high recall detection performance. Unlike a binary classifier, a binary detection model has a limited absolute recall level, which is often below 100\%. We aim for this level of absolute recall to be as high as possible and improve the precision at those recall levels by use of the subsequent classifier.
Based on our extensive experiments, our \textit{broad-to-narrow} model achieves both an overall improvement (average precision score) over end-to-end models, and an improvement in precision at the high recall end. 

To achieve improved high recall detection, we split the problem into coarse localization and fine grained classification. 
We pick two general object classes and in both cases, we possess a modest number (on the order of 1,000-10,000, see section \ref{datasets} for more details) of ground truth annotations on high resolution satellite imagery. 
However, in some of these sub-classes we have considerably less training data, so synthetic data becomes more valuable.

Since our problems have a visually coherent class hierarchy, they are good candidates for our two stage \textit{broad-to-narrow} classification/detection system.
This algorithm works by first learning a one class, general detection task, then learning to classify the detects of the first stage.
The detection stage is trained to detect not just the target sub-class but rather all members of the parent class on an equal basis.
Once the detection model has been trained, it is applied to test images and 64x64 pixel crops, concentric on predicted boxes, are collected to be forwarded to the subsequent classifier.

Separately, we train models to classify detected objects into the fine grained sub-categories.
This is done using the same training set annotations used to train the detector, but centered and cropped to 64x64 pixel size.
We experimented with several classification models. Our first attempts were based on ResNet-50 feature extraction, followed by either a logistic regression, a small dense neural network, or a random forest. We found these classifier models led to mediocre \textit{broad-to-narrow} models which were on par with or slightly worse than the corresponding end-to-end detection models. This is perhaps unsurprising since the internal classification branch of Faster R-CNN is a similar low capacity model.

Our second attempt was much more powerful: we trained ResNet models end-to-end (pre-trained from Torchvision \cite{torchvision}) on the fine grained classification task. With these higher capacity models we were also able to use data augmentation (rotations, shifts, noise, and color jitter) to mitigate the overfitting of the ResNet to our small training data sets. The resulting \textit{broad-to-narrow} models out-perform the end-to-end models.

\subsection{Scoring strategy for Broad-to-Narrow}
\begin{figure}
    \centering
    \includegraphics[width=15em]{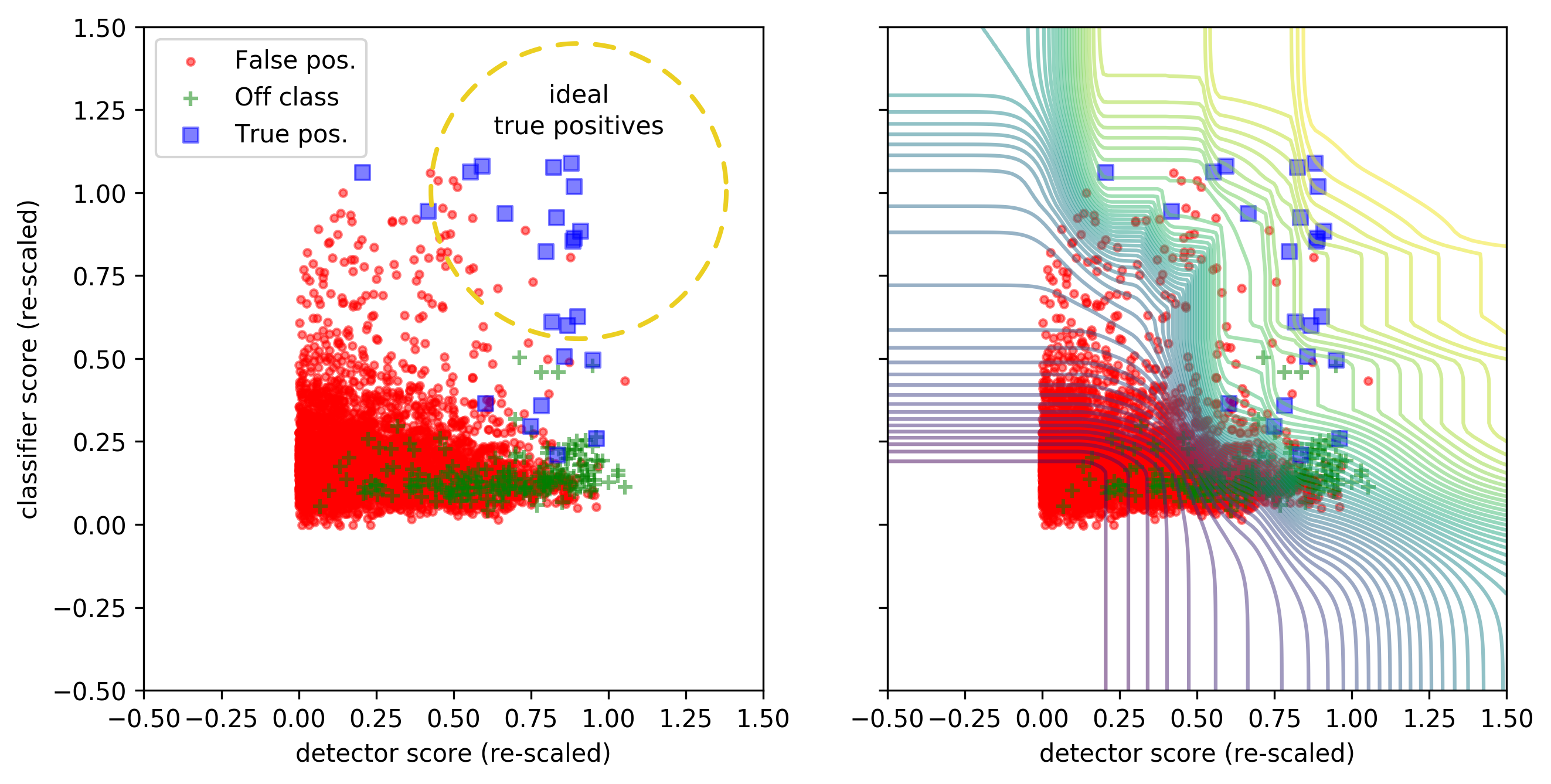}
    \caption{Left\textemdash an example of the (validation set) scores coming from a two stage model showing scores assigned to false detects, true general detects not of class A, and true detects of the target sub-class.
    Right\textemdash contours of the modified KDE used to ensemble the detector and classifier scores.
    }
    \label{twostagescatter}
\end{figure}

Having trained detector and classifier, we must next decide how to interpret the combination of signals from these two models.
Given ideal detector and classifier, we could separate the false detects from true general detects by the detector score, and further separate our sub-class using the classifier score. In reality, the model scores are distributed more like Fig. \ref{twostagescatter}. Like a precision/recall curve for a single score model, we wish to capture the trade-off between recall and precision offered by our two stage model. 

One approach is to traverse the range of confidence thresholds for the two component models independently. Given confidence thresholds for the detector and classifier, decide a detect box is acceptable if it surpasses the detector threshold and the classifier threshold.
By sweeping detector and classifier confidence through their full ranges, we can record an array of different precision and recall values. Though similar in spirit to a precision-recall curve, this summary of model performance has two free parameters, so it unfairly over-represents the performance of the two stage model. Considering the scatter plot in Fig. \ref{twostagescatter}, picking thresholds for each detector and classifier results in a decision boundary equal to the boundary of an upper right quadrant of the scatter plot. The method described above corresponds to recording and plotting the precision and recall from every possible such upper-right quadrant boundary.

To make a fair comparison to end-to-end models, we need to somehow reduce the detector and classifier scores to one measure.
One possibility is to fix a threshold for the classifier and sweep the detector threshold through a range, plotting precision and recall. 
This would be reasonable given a near ideal classifier because a single decision boundary could be set for the fine grained classification task with near zero error. Another possibility is to set a single threshold for the detector and vary the classifier confidence; for the same reason this would work well for a near perfect detector.

However, our \textit{broad-to-narrow} models are composed of detectors and classifiers which are imperfect, and we find the best total performance by combining their confidences into one ensemble score.
To do so we use the validation set to construct a density estimate (via the Gaussian kernel density estimator) of the negative class in the two-dimensional (detector score,classifier score) space. 
Let $s_D$ and $s_C$ be the pre-sigmoid classifier and detector scores, normalized to the interval $[0,1]$ by a linear transformation.\footnote{The normalization is fitted using negative class validation set samples.}
These two scores we can consider coordinate functions on detects, so every detect\textemdash false positive, off-class detect, true positive\textemdash is located in two dimensional space, in or near to the unit square $[0,1]\times [0,1]$. We form a KDE of the negative samples (false detects and off-class true detects), let us call it $\mathcal{N}$, and a KDE of the positive class, called $\mathcal{P}$. In the zero-shot setting, our validation set contains no target class samples, and we substitute $\mathcal{P} = \mbox{min}(s_D,s_C)$. To represent $s_D$ and $s_C$ as model confidence scores in the KDE's, we set
\[\overline{\mathcal{N}}(x,y) = \max_{s\geq x, t\geq y}\left[\mathcal{N}(s,t)\right]\]
This enforces the following condition: if A and B are two detect boxes, and both of B's score are lower than A's, then B is at least as likely to be a negative example as A (we estimate). Likewise let
\[\overline{\mathcal{P}}(x,y) = \max_{s\leq x, t\leq y}\left[\mathcal{P}(s,t)\right]\]
Finally, the ensemble score for detector and classifier $D$ and $C$ and detect $x$ is 
\[\mathcal{E}_{D,C}(x) = \overline{\mathcal{P}}(s_D(x),s_C(x)) - \overline{\mathcal{N}}(s_D(x),s_C(x))\]
We found this to be an effective way to combine the knowledge of the detector and classifier within our two stage model.

\subsection{CAD based synthetics}
As described in the previous section, 3D CAD models as training data have shown promise in the literature. We have been able to replicate these successes on some problems, so we included this conventional method as a component in our synthetic data generator mixture. 

Our processes for creating synthetic training images from 3D CAD models are inspired by the goals of domain matching and domain randomization. 
In line with domain matching, when we integrate CAD models onto background images, we avoid letting the synthetic object ``stand out'' from the background.
To achieve this level of blending, we apply approximate histogram matching by adjusting the saturation and value levels to the background. When encountering image attributes that are intractable or difficult to match, we employ domain randomization. For example, we superimpose aircraft with random camouflage skins and onto random backgrounds taken over airfields.

Our process starts by acquiring one or more realistic 3D polygon mesh models of an object of interest. This object is then rendered with neutral gray skin and in various different poses (off nadir angle, look angle, sun angle) on a transparent background, but a semi-transparent shadow. Next, we select reasonable background satellite imagery from our library and alpha blend the 3D renders onto these background chips. At this point in the process we may use ground truth annotations to avoid placing an object of interest on top of some actual objects (for instance, no rail cars are pasted on top of buildings) and at the very least we guarantee that there are no real examples from the target class in the background. In the case of our rail car 3D models, we found it essential to arrange the rail cars in rows, just as they are in real imagery.

\begin{figure}
	\centering
  	\includegraphics[width=\columnwidth]{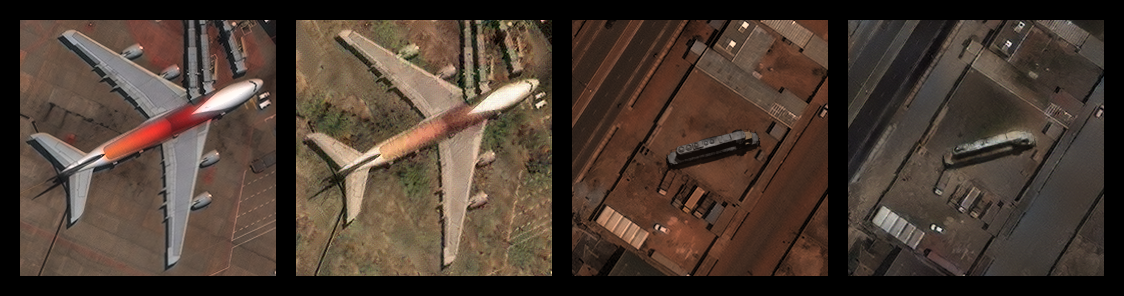}
  	\caption{Two synthetic data techniques\textemdash neural style transfer and GAN reskinning\textemdash tested in this paper.}
  	\Description{Two synthetic data techniques\textemdash neural style transfer and GAN reskinning\textemdash tested in this paper.}
  	\label{fig:teaser}
\end{figure}

Once a 3D model has been placed with a particular pose on a given background image, the statistics of the local portion of the background are used to inform the coloring of the 3D model. A paint pattern such as Gaussian noise or military camouflage may be applied at this stage, and the saturation, value, etc. of the local portion of the background are taken into account to modify the paint. As post processing, some small blur is applied to dull the sharp contrast between the 3D model and the background. For an example of our 3D synthetics, see the third tile of Fig. \ref{fig:teaser}.

\subsection{GAN-Reskinner synthetics}

\begin{figure}
    \centering
    \includegraphics[width=15em]{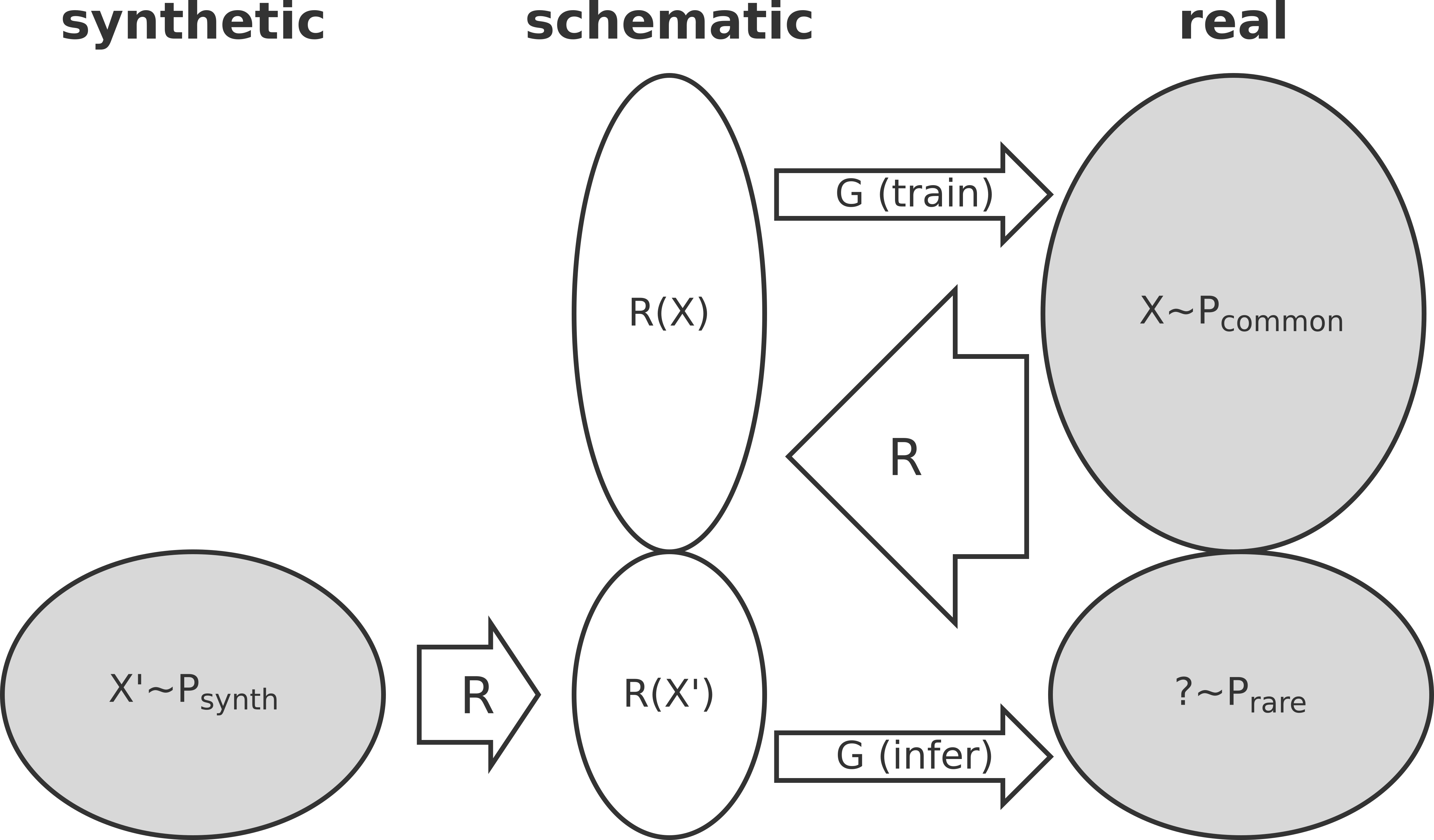}
    \caption{Diagram of the GAN reskinning system.}
    \label{gan_diagram}
\end{figure}

Pasting 3D CAD models onto satellite image chips can provide effective training data for a detection model.
One of the major challenges of this method, though, is to sufficiently blend the 3D models into the background.
The concern is that detectors may overfit to the relatively easy problem of detecting 3D models if those synthetic objects are highly distinguishable from real objects and easily separated from the background. The detectors as a result will fail when tested on more challenging examples from the real data distribution. Informally, 3D rendered models often seem too ``clean'' and the data's training utility might be improved by perturbing them in some appropriate way.

To address this problem, we extend the ``pix2pix'' \cite{pix2pix} GAN to build a reskinning system for our 3D modeled detection scenes (see Fig. \ref{gan_diagram}). One of the basic insights from pix2pix is that a GAN can be taught to rebuild an image, with passable accuracy and realism, from a figurative or schematic representation. Our idea starts with finding an automated way to compute this schematic representation $R(X)$ of a high resolution satellite image $X$ (e.g. edge detects of some kind). The function $R$ should sufficiently modify $X$ that it is difficult to decide, based on $R(X)$ alone, whether $X$ was a real satellite image or one synthesized from 3D models, but the outputs of $R$ should also be detailed enough that the pix2pix generator can use them to reconstruct good looking satellite imagery.

\begin{figure}
    \centering
    \includegraphics[width=\columnwidth]{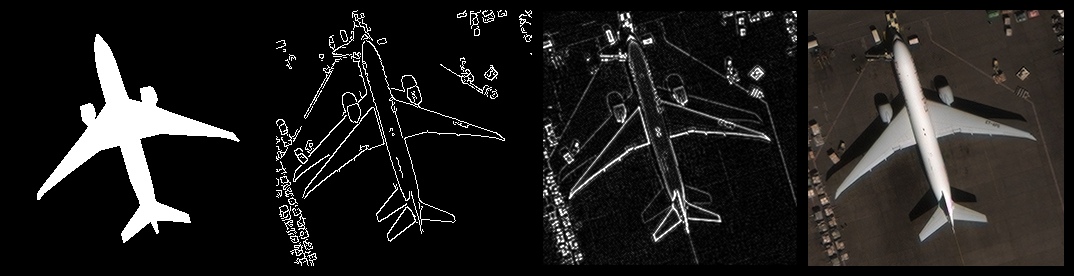}
    \caption{Different types of schematic representation functions $R$ tested for training pix2pix.}
    \label{edges}
\end{figure}

Once the representation function $R$ has been chosen, a pix2pix GAN is trained to approximate the conditional distributions \newline $(X|R(X)=R_0)$, 
i.e. the generator $G$ should turn representations back into images so that $G(R(X))$ is a likely input to $R$.

After these two components are integrated, the GAN-Reskinner system will take a synthetic scene $X'$, which includes objects of some rare class, and return $G(R(X'))$.
The pix2pix GAN first learns the relationship between $X\sim P_{\mbox{common}}$ and corresponding $R(X)$. In the inference stage, we leverage the generator's knowledge of this relationship to reconstruct a passable real counterpart of $R(X')$.

In order of increasing detail preserved, the representations we tried were segmentation polygons, Canny edge detects, and edge detects from a simple 3x3 Laplacian kernel operator (see Fig. \ref{edges}). We trained pix2pix networks for all three choices of $R$, but we only tested the latter in the application of training data improvement.
In addition to picking the representation function $R$, there is the choice of what pairs $(X,R(X))$ to show to the GAN.
Suppose we wish to use the reskinning system to treat scenes in which 3D aircraft models have been embedded.
One choice would be to show the GAN pairs $(X,R(X))$ where $X$ comes from a distribution of real airport images, including some more common types of aircraft.
Another choice, which we ultimately picked, is to show the GAN a wide range of satellite imagery including buildings, ground vehicles, etc.
In particular, our goal is to test the reskinning system described above when the GAN's training set contains no real images of the target class. 
An example output is shown in the fourth tile of Fig. \ref{fig:teaser}.

\subsection{Neural style transfer augmentation}
\label{nnstxaug}

In general, the satellite images on which we wish to run our models can represent many different regions, seasons, lighting conditions, etc. The resulting differences in image qualities can lead to poor model generalization, but these qualities are difficult to capture quantitatively. To address this problem, we use neural style transfer \cite{nnstx} as a data augmentation technique. By codifying ``image qualities'' as the Gram matrices of deep features which drive neural style transfer, such augmentation will lead to detection models that are less sensitive to particular satellite imaging conditions and hence will generalize better.

To this end, we select $k$ style images which we use to multiply any data set of real training images $k$-fold. We primarily follow \cite{nnstx} with the modification that we use the ResNet-50 network, in place of VGG-19, to capture deep features. Prior to neural style transfer, we use linear histogram matching (based on Mahalanobis whitening) to give the color palette of the content image a similar distribution to that of the style image. 
Style targets are taken from the Gram matrices of feature maps before some initial segment of the four meta-blocks, and the content target is taken from the features after the last of these meta-blocks.
The style transfer is initialized with the color transformed content image and L-BFGS optimization is used to minimize the sum of the losses.
See the second tile of Figure \ref{fig:teaser} for an example.

\section{Results}
\label{results}

\subsection{Overview of experiments}

We performed rigorous and comprehensive tests to evaluate our methods on multiple target classes, to be detailed fully in the next section.
We conducted two sets of trials for each target class:
\begin{enumerate}
    \item We tested the efficacy of 3D CAD models, GAN reskinned models, neural styled real data, and mixtures thereof, in the end-to-end detector, low and zero-sample settings. In the zero-sample setting, we do not include neural style transfer as we have applied that method to real data.
    \item We tested our \textit{broad-to-narrow} architecture in the low-sample and zero-sample settings, using the data sources described above to train the detectors and using 3D CAD models, GAN reskinned models, and neural style transfer applied to real and synthetic images to train the classifiers.
\end{enumerate}
Having conducted these extensive tests, we are able to glean more than just a proof of concept for these methods\textemdash we can also observe patterns in performance which lead to important insights and future work in synthetic data utility.

\subsection{Data sets}
\label{datasets}

To evaluate our methods of creating synthetic data, we trained models targeting two application domains: aircraft detection and classification and rail car detection and classification. These target domains give us an array of sub-variants to which we can focus our attention. In addition to detecting a rail car or plane, we will classify it as one of a few sub-classes of these two parent classes.

Besides testing our methods on low sample problems, we are also interested in their impact on the zero sample problem.
To simulate the zero sample problem, we simply prohibit ourselves from using any real training or validation examples of sub-class X when developing a model for testing against sub-class X.
Synthetic examples from sub-class X are allowed, but neural style transfer done on a real image containing sub-class X is not.

Both the aircraft data set and the rail data set were composed of 30cm, rgb satellite imagery from the WorldView-3 sensor. Post processing included dynamic range adjustment, atmospheric compensation, and pan-sharpening.
Including some data from the xView data set \cite{xview}, all images were annotated with bounding boxes and class labels.
For both data sets, train/val/test splits were made based on geographic location, i.e. no training image is taken from at or near the same location as a test image.

The aircraft data set is classified into two target sub-classes (type A and type B) as well as third `miscellaneous' sub-class.
The images were cropped into chips of size 768x768 (or around 230x230 meters) for the purposes of feeding to the detection model. 
The test set for aircraft detection totals 1646 mega pixels of 30cm satellite imagery; there are 1103, 256, and 274 objects in the train, validation, and test sets, respectively.
The rail data set is classified into seven classes (cargo, flat, general, high speed, locomotive, passenger, and tanker). The passenger car and tanker car classes we selected as the target of our experiments.
The images were cropped to a smaller 384x384 px. to avoid having too many objects in one chip.
The rail test set is comprised of 1248 mega pixels of 30cm satellite imagery; there are 8438, 4085, and 3050 objects labelled in the train, validation, and test sets, respectively.

We did not test every possible synthetic data combination, due to computational constraints. For zero-shot detection we include 3D CAD, GAN-Reskinner, and combinations, with and without neural style transfer applied. For low-shot detection we use real, 3D CAD, GAN-Reskinner, and neural style transfer applied to 3D CAD data, and combinations.

When reporting model performance trained on various combinations of synthetic data, we use the codes: R = real, C = 3D CAD, G = GAN reskinned, N = neural style transfer. In some cases we apply neural style transfer to synthetic data (CAD and GAN), and thus treating neural style transfer as a post-processing step, write X.plain, X.nnstx, X.both for no post-processing, neural style transfer post-processing, or both applied to data source X.

\subsection{Scoring}
\label{results_sec}
To measure the performance of the classifiers by themselves, we report their performance on the classification test set (simulating a perfect detector). We plot the ROC curve and prefer to measure the percent area \textit{over} the curve (lower is better, random classifier scores 50\%). See Table \ref{roc_aoc} for these numbers.

\begin{figure}[ht]
    \centering
    \includegraphics[width=\columnwidth]{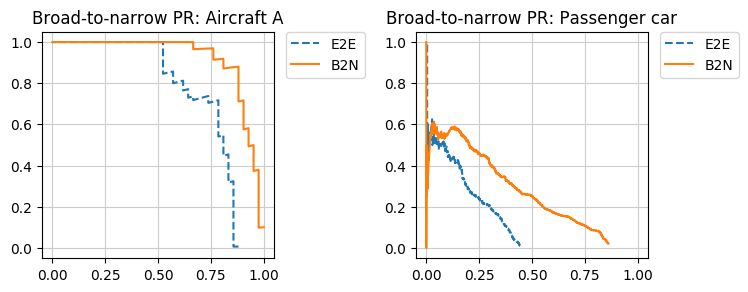}
    \caption{Examples of zero-shot precision/recall curves.
    }
    \label{PR_example}
\end{figure}

We score detectors in the usual fashion with one added prepossessing step. Rather than matching model predictions to ground truth boxes in pixel space (and per-chip), we first convert all predictions into geospatial coordinates, use non-maximum suppression to stitch them together, and then perform matching as usual.
It is also worth noting that we run inference over all the image chips from a large region of interest, not just those chips that actually contain objects.
Following typical scoring methods for object detection tasks, we match predicted boxes with ground truth boxes based on the boxes' IoU. 
Prediction boxes with highest detector confidence are given preference, and match to the ground truth box with which they have highest IoU of at least 50\%. 
In particular, duplicate predictions at one ground truth box are counted as false positives.
Based on this matching, along with the predictions' confidence scores, we compute several statistics: precision and recall at any threshold
(see figure \ref{PR_example} for examples)
and average precision (AP50) as a summary statistic. 

\subsection{Synthetic data evaluation}

We tested several mixtures of real and synthetic data for both the broad detection problems (all aircraft, all rail cars) and the fine grained problems. These simulate low sample problems.
We also created datasets with only synthetic objects, simulating a zero-shot scenario, and tested these for training fine grained detection models.
We first evaluate the utility of these data sources for training traditional classifiers and detectors, before moving on to evaluating the \textit{broad-to-narrow} models in the next section.

The AP50 scores for the detection runs are reported in Table \ref{alle2eresults}.
In every zero-shot case but one, we see an improvement in AP score by including GAN reskinned synthetics in the training of the detector.
In the only zero-shot case where GAN didn't improve performance, all end-to-end detectors failed to produce any appreciable performance (0.0\% AP).
The low-shot cases are more mixed, but in eight out of twelve trials, including GAN synthetics improved AP.
In pure classification tasks, see Table \ref{roc_aoc}, the situation is also mixed, with no clear evidence in favor of the GAN-Reskinner.
In low/zero-shot classification and detection, model response to neural style transfer synthetics is mixed.
Although model response to synthetic data seems to be highly class dependent, including GAN synthetics alongside regular CAD synthetics seems to provide a consistent benefit in the zero-shot detection case.

\begin{table}[h]
	\tiny
    \centering
    \caption{End-to-end detection AP scores, aircraft and rail.}
	\begin{tabular}[t]{|l|rrr|}
        \hline
        \backslashbox{detr.}{class} & General & Type A & Type B \\
        \hline
        \multicolumn{4}{|c|}{Low-shot}\\
        \hline
        R & 80.9\% & 38.2\% & 62.7\% \\
        R+C & 94.1\% & 71.0\% & 82.7\% \\
        R+C+G & 93.9\% & 72.5\% & 83.6\% \\
        R+N & 87.6\% & 63.2\% & 67.3\% \\
        R+C+N & 94.4\% & 64.5\% & 84.5\% \\
        R+C+G+N & \textbf{96.7}\% & 67.5\% & \textbf{85.5\%} \\
        \hline
        \multicolumn{4}{|c|}{Zero-shot}\\
        \hline
        C &  & 32.6\% & 30.7\% \\
        G &  & 70.4\% & 34.5\% \\
        C+G &  & \textbf{75.7\%} & \textbf{41.1\%} \\
        \hline
    \end{tabular}
	\begin{tabular}[t]{|l|rrr|}
        \hline
        \backslashbox{detr.}{class} & All rail & Passenger & Tanker \\
        \hline
        \multicolumn{4}{|c|}{Low-shot}\\
        \hline
        R & 39.2\% & 77.6\% & 39.7\% \\
        R+C & 41.3\% & \textbf{78.3\%} & 47.9\% \\
        R+C+G & \textbf{41.5\%} & 76.5\% & 46.7\% \\
        R+N & 40.2\% & 73.1\% & 41.7\% \\
        R+C+N & 39.4\% & 77.4\% & 46.2\% \\
        R+C+G+N & 41.4\% & 69.4\% & \textbf{49.7\%} \\
        \hline
        \multicolumn{4}{|c|}{Zero-shot}\\
        \hline
        C &  & 12.0\% & 0.0\% \\
        G &  & 6.4\% & 0.0\% \\
        C+G &  & \textbf{13.3\%} & 0.0\% \\
        \hline
    \end{tabular}
    \label{alle2eresults}
\end{table}

\begin{table}[h]
	\tiny
    \centering
    \caption{Area over ROC curve, low and zero-shot classifiers.}
    \begin{tabular}{|l|rrrr|}
        \hline
        \backslashbox{clsr.}{class} &  Aircraft A &  Aircraft B &  Passenger car &  Tanker car \\
        \hline
        \multicolumn{5}{|c|}{Low-shot}\\
        \hline
        R          &\cellcolor[RGB]{255,255,255}      0.00\% &\cellcolor[RGB]{223,255,223}      3.81\% &\cellcolor[RGB]{243,255,243}          1.41\% &\cellcolor[RGB]{176,255,176}       9.29\% \\
        R+N        &\cellcolor[RGB]{255,255,255}      0.00\% &\cellcolor[RGB]{244,255,244}      1.32\% &\cellcolor[RGB]{245,255,245}          1.18\% &\cellcolor[RGB]{179,255,179}       8.99\% \\
        R+G        &\cellcolor[RGB]{255,255,255}      0.00\% &\cellcolor[RGB]{241,255,241}      1.60\% &\cellcolor[RGB]{247,255,247}          0.98\% &\cellcolor[RGB]{203,255,203}       6.19\% \\
        R+G+N      &\cellcolor[RGB]{255,255,255}      0.00\% &\cellcolor[RGB]{241,255,241}      1.71\% &\cellcolor[RGB]{245,255,245}          1.18\% &\cellcolor[RGB]{192,255,192}       7.49\% \\
        R+C        &\cellcolor[RGB]{255,255,255}      0.00\% &\cellcolor[RGB]{245,255,245}      1.21\% &\cellcolor[RGB]{250,255,250}          0.58\% &\cellcolor[RGB]{155,255,155}      11.80\% \\
        R+C+N      &\cellcolor[RGB]{255,255,255}      0.00\% &\cellcolor[RGB]{225,255,225}      3.56\% &\cellcolor[RGB]{246,255,246}          1.08\% &\cellcolor[RGB]{180,255,180}       8.87\% \\
        R+C+G      &\cellcolor[RGB]{255,255,255}      0.00\% &\cellcolor[RGB]{233,255,233}      2.57\% &\cellcolor[RGB]{245,255,245}          1.23\% &\cellcolor[RGB]{199,255,199}       6.64\% \\
        R+C+G+N    &\cellcolor[RGB]{255,255,255}      0.00\% &\cellcolor[RGB]{243,255,243}      1.41\% &\cellcolor[RGB]{247,255,247}          0.89\% &\cellcolor[RGB]{184,255,184}       8.35\% \\
        \hline
        \multicolumn{5}{|c|}{Zero-shot}\\
        \hline
        C.plain    &\cellcolor[RGB]{254,255,254}      0.28\% &\cellcolor[RGB]{163,255,163}     19.28\% &\cellcolor[RGB]{240,255,240}          3.06\% &\cellcolor[RGB]{179,255,179}      15.84\% \\
        C.both     &\cellcolor[RGB]{254,255,254}      0.18\% &\cellcolor[RGB]{177,255,177}     16.35\% &\cellcolor[RGB]{231,255,231}          4.97\% &\cellcolor[RGB]{191,255,191}      13.31\% \\
        C.nnstx    &\cellcolor[RGB]{253,255,253}      0.45\% &\cellcolor[RGB]{159,255,159}     19.98\% &\cellcolor[RGB]{228,255,228}          5.65\% &\cellcolor[RGB]{198,255,198}      11.85\% \\
        C+G.plain  &\cellcolor[RGB]{255,255,255}      0.06\% &\cellcolor[RGB]{155,255,155}     20.84\% &\cellcolor[RGB]{224,255,224}          6.49\% &\cellcolor[RGB]{201,255,201}      11.24\% \\
        C+G.both   &\cellcolor[RGB]{254,255,254}      0.30\% &\cellcolor[RGB]{171,255,171}     17.49\% &\cellcolor[RGB]{222,255,222}          6.83\% &\cellcolor[RGB]{191,255,191}      13.46\% \\
        C+G.nnstx  &\cellcolor[RGB]{252,255,252}      0.67\% &\cellcolor[RGB]{168,255,168}     18.14\% &\cellcolor[RGB]{228,255,228}          5.73\% &\cellcolor[RGB]{193,255,193}      12.91\% \\
        G.plain    &\cellcolor[RGB]{251,255,251}      0.91\% &\cellcolor[RGB]{167,255,167}     18.37\% &\cellcolor[RGB]{235,255,235}          4.21\% &\cellcolor[RGB]{165,255,165}      18.72\% \\
        G.both     &\cellcolor[RGB]{248,255,248}      1.48\% &\cellcolor[RGB]{167,255,167}     18.45\% &\cellcolor[RGB]{213,255,213}          8.68\% &\cellcolor[RGB]{187,255,187}      14.27\% \\
        G.nnstx    &\cellcolor[RGB]{250,255,250}      1.08\% &\cellcolor[RGB]{183,255,183}     15.13\% &\cellcolor[RGB]{233,255,233}          4.53\% &\cellcolor[RGB]{168,255,168}      18.18\% \\
        \hline
    \end{tabular}
    \label{roc_aoc}
\end{table}

\subsection{Broad-to-narrow evaluation}

In tables \ref{b2n_lo_shot_results} and \ref{b2n_zero_shot_results} we have compiled the AP scores from our various \textit{broad-to-narrow} models. Each column is a detector trained on a different mixture of synthetic and real training data, and each row is a different classifier. 
The classifiers are finetuned ResNet-50 models, trained on some mixture of 3D CAD and GAN synthetic data.
Finetuning
made it possible to use robust data augmentation techniques from standard computer vision practice and hence add a lot of value to the \textit{broad-to-narrow} model.

For every broad stage detector, we compare the resulting two stage models to a similarly trained end-to-end model.
We compare low-shot models to low-shot models, and zero-shot models to other zero-shot models.
In the zero-shot case, the end-to-end models are developed with no real images in their training sets.
The end-to-end models trained for our two aircraft sub-classes are able to learn well by only seeing synthetic objects,
but in the case of tanker cars, the end-to-end model learns virtually nothing transferable from synthetic data alone.
However, by training with sibling class rail cars at the detection stage and using synthetics to train the classifier, we achieve good performance on that class with our \textit{broad-to-narrow} strategy.
In the low-shot case, the difference is less stark, but \textit{broad-to-narrow} still outperforms the end-to-end models.
In every case, our \textit{broad-to-narrow} models beat their corresponding end-to-end model, both in low and zero-shot settings.

Unlike the zero-shot end-to-end detection experiment, in which the GAN-Reskinner proved useful, we do not see consistent, patterns of synthetic data utility across classes. 
Nonetheless, having more varieties of synthetic data available for training \textit{broad-to-narrow} models usually allows us to unlock higher performance, though at the cost of an extra hyper-parameter search.

\begin{table}
    \centering
    \caption{AP scores broad-to-narrow vs end-to-end detection, low-shot.}
    \tiny
    \begin{tabular}{|l|rrrrrr|}
        \hline
        \multicolumn{7}{|c|}{Aircraft A}\\
        \hline
        \backslashbox{clsr.}{detr.} &     R &   R+C &  R+C+G &   R+N &  R+C+N &  R+C+G+N \\
        \hline
        End-to-End &\cellcolor[RGB]{255,255,255} 38.2\% &\cellcolor[RGB]{252,255,252} 71.0\% &\cellcolor[RGB]{246,255,246}  72.5\% &\cellcolor[RGB]{255,255,255} 63.2\% &\cellcolor[RGB]{255,255,255}  64.5\% &\cellcolor[RGB]{255,255,255}    67.5\% \\
        R          &\cellcolor[RGB]{239,255,239} 74.7\% &\cellcolor[RGB]{178,255,178} 92.5\% &\cellcolor[RGB]{181,255,181}  91.8\% &\cellcolor[RGB]{202,255,202} 85.5\% &\cellcolor[RGB]{161,255,161}  97.6\% &\cellcolor[RGB]{159,255,159}    98.0\% \\
        R+N        &\cellcolor[RGB]{242,255,242} 73.7\% &\cellcolor[RGB]{182,255,182} 91.5\% &\cellcolor[RGB]{170,255,170}  94.9\% &\cellcolor[RGB]{189,255,189} 89.3\% &\cellcolor[RGB]{161,255,161}  97.6\% &\cellcolor[RGB]{158,255,158}    98.4\% \\
        R+G        &\cellcolor[RGB]{239,255,239} 74.6\% &\cellcolor[RGB]{173,255,173} 94.1\% &\cellcolor[RGB]{164,255,164}  96.7\% &\cellcolor[RGB]{189,255,189} 89.4\% &\cellcolor[RGB]{161,255,161}  97.6\% &\cellcolor[RGB]{157,255,157}    98.8\% \\
        R+G+N      &\cellcolor[RGB]{238,255,238} 74.9\% &\cellcolor[RGB]{172,255,172} 94.3\% &\cellcolor[RGB]{162,255,162}  97.2\% &\cellcolor[RGB]{183,255,183} 91.0\% &\cellcolor[RGB]{161,255,161}  97.6\% &\cellcolor[RGB]{157,255,157}    98.6\% \\
        R+C        &\cellcolor[RGB]{239,255,239} 74.7\% &\cellcolor[RGB]{173,255,173} 94.1\% &\cellcolor[RGB]{163,255,163}  97.1\% &\cellcolor[RGB]{189,255,189} 89.2\% &\cellcolor[RGB]{161,255,161}  97.6\% &\cellcolor[RGB]{163,255,163}    97.0\% \\
        R+C+N      &\cellcolor[RGB]{242,255,242} 73.9\% &\cellcolor[RGB]{180,255,180} 92.0\% &\cellcolor[RGB]{169,255,169}  95.3\% &\cellcolor[RGB]{194,255,194} 88.0\% &\cellcolor[RGB]{161,255,161}  97.6\% &\cellcolor[RGB]{167,255,167}    95.9\% \\
        R+C+G      &\cellcolor[RGB]{239,255,239} 74.7\% &\cellcolor[RGB]{170,255,170} 94.9\% &\cellcolor[RGB]{158,255,158}  98.5\% &\cellcolor[RGB]{186,255,186} 90.3\% &\cellcolor[RGB]{161,255,161}  97.6\% &\cellcolor[RGB]{155,255,155}    \textbf{99.3\%} \\
        R+C+G+N    &\cellcolor[RGB]{241,255,241} 74.0\% &\cellcolor[RGB]{175,255,175} 93.4\% &\cellcolor[RGB]{161,255,161}  97.6\% &\cellcolor[RGB]{188,255,188} 89.7\% &\cellcolor[RGB]{161,255,161}  97.6\% &\cellcolor[RGB]{158,255,158}    98.3\% \\
        \hline
        \multicolumn{7}{|c|}{Aircraft B}\\
        \hline
        \backslashbox{clsr.}{detr.} &     R &   R+C &  R+C+G &   R+N &  R+C+N &  R+C+G+N \\
        \hline
        End-to-End &\cellcolor[RGB]{255,255,255} 62.9\% &\cellcolor[RGB]{238,255,238} 82.7\% &\cellcolor[RGB]{234,255,234}  83.4\% &\cellcolor[RGB]{255,255,255} 67.3\% &\cellcolor[RGB]{230,255,230}  84.0\% &\cellcolor[RGB]{223,255,223}    85.1\% \\
        R          &\cellcolor[RGB]{192,255,192} 90.2\% &\cellcolor[RGB]{169,255,169} 93.9\% &\cellcolor[RGB]{174,255,174}  93.0\% &\cellcolor[RGB]{202,255,202} 88.6\% &\cellcolor[RGB]{179,255,179}  92.2\% &\cellcolor[RGB]{184,255,184}    91.4\% \\
        R+N        &\cellcolor[RGB]{213,255,213} 86.7\% &\cellcolor[RGB]{177,255,177} 92.5\% &\cellcolor[RGB]{188,255,188}  90.8\% &\cellcolor[RGB]{199,255,199} 89.0\% &\cellcolor[RGB]{162,255,162}  95.0\% &\cellcolor[RGB]{155,255,155}    \textbf{96.1\%} \\
        R+G        &\cellcolor[RGB]{200,255,200} 88.9\% &\cellcolor[RGB]{167,255,167} 94.2\% &\cellcolor[RGB]{182,255,182}  91.7\% &\cellcolor[RGB]{188,255,188} 90.8\% &\cellcolor[RGB]{157,255,157}  95.8\% &\cellcolor[RGB]{159,255,159}    95.5\% \\
        R+G+N      &\cellcolor[RGB]{211,255,211} 87.1\% &\cellcolor[RGB]{175,255,175} 92.8\% &\cellcolor[RGB]{193,255,193}  90.0\% &\cellcolor[RGB]{198,255,198} 89.1\% &\cellcolor[RGB]{164,255,164}  94.7\% &\cellcolor[RGB]{162,255,162}    95.0\% \\
        R+C        &\cellcolor[RGB]{192,255,192} 90.1\% &\cellcolor[RGB]{166,255,166} 94.4\% &\cellcolor[RGB]{174,255,174}  93.0\% &\cellcolor[RGB]{189,255,189} 90.6\% &\cellcolor[RGB]{159,255,159}  95.4\% &\cellcolor[RGB]{157,255,157}    95.8\% \\
        R+C+N      &\cellcolor[RGB]{225,255,225} 84.9\% &\cellcolor[RGB]{192,255,192} 90.2\% &\cellcolor[RGB]{200,255,200}  88.8\% &\cellcolor[RGB]{210,255,210} 87.2\% &\cellcolor[RGB]{174,255,174}  93.0\% &\cellcolor[RGB]{173,255,173}    93.2\% \\
        R+C+G      &\cellcolor[RGB]{209,255,209} 87.4\% &\cellcolor[RGB]{175,255,175} 92.8\% &\cellcolor[RGB]{192,255,192}  90.1\% &\cellcolor[RGB]{215,255,215} 86.5\% &\cellcolor[RGB]{169,255,169}  93.9\% &\cellcolor[RGB]{166,255,166}    94.3\% \\
        R+C+G+N    &\cellcolor[RGB]{206,255,206} 87.9\% &\cellcolor[RGB]{170,255,170} 93.7\% &\cellcolor[RGB]{185,255,185}  91.3\% &\cellcolor[RGB]{202,255,202} 88.6\% &\cellcolor[RGB]{162,255,162}  94.9\% &\cellcolor[RGB]{159,255,159}    95.4\% \\
        \hline
        \multicolumn{7}{|c|}{Passenger rail}\\
        \hline
        \backslashbox{clsr.}{detr.} &     R &   R+C &  R+C+G &   R+N &  R+C+N &  R+C+G+N \\
        \hline
        End-to-End &\cellcolor[RGB]{255,255,255} 77.5\% &\cellcolor[RGB]{255,255,255} 78.2\% &\cellcolor[RGB]{255,255,255}  76.4\% &\cellcolor[RGB]{255,255,255} 72.9\% &\cellcolor[RGB]{255,255,255}  77.2\% &\cellcolor[RGB]{255,255,255}    69.2\% \\
        R          &\cellcolor[RGB]{211,255,211} 83.2\% &\cellcolor[RGB]{213,255,213} 83.1\% &\cellcolor[RGB]{255,255,255}  78.1\% &\cellcolor[RGB]{182,255,182} 85.3\% &\cellcolor[RGB]{193,255,193}  84.5\% &\cellcolor[RGB]{218,255,218}    82.7\% \\
        R+N        &\cellcolor[RGB]{228,255,228} 82.0\% &\cellcolor[RGB]{210,255,210} 83.3\% &\cellcolor[RGB]{248,255,248}  80.5\% &\cellcolor[RGB]{187,255,187} 85.0\% &\cellcolor[RGB]{158,255,158}  87.1\% &\cellcolor[RGB]{202,255,202}    83.9\% \\
        R+G        &\cellcolor[RGB]{200,255,200} 84.0\% &\cellcolor[RGB]{214,255,214} 83.0\% &\cellcolor[RGB]{240,255,240}  81.1\% &\cellcolor[RGB]{180,255,180} 85.5\% &\cellcolor[RGB]{171,255,171}  86.1\% &\cellcolor[RGB]{203,255,203}    83.8\% \\
        R+G+N      &\cellcolor[RGB]{241,255,241} 81.0\% &\cellcolor[RGB]{193,255,193} 84.5\% &\cellcolor[RGB]{239,255,239}  81.2\% &\cellcolor[RGB]{203,255,203} 83.8\% &\cellcolor[RGB]{182,255,182}  85.3\% &\cellcolor[RGB]{211,255,211}    83.2\% \\
        R+C        &\cellcolor[RGB]{207,255,207} 83.5\% &\cellcolor[RGB]{189,255,189} 84.8\% &\cellcolor[RGB]{214,255,214}  83.0\% &\cellcolor[RGB]{182,255,182} 85.3\% &\cellcolor[RGB]{156,255,156}  87.2\% &\cellcolor[RGB]{185,255,185}    85.1\% \\
        R+C+N      &\cellcolor[RGB]{218,255,218} 82.7\% &\cellcolor[RGB]{214,255,214} 83.0\% &\cellcolor[RGB]{226,255,226}  82.1\% &\cellcolor[RGB]{187,255,187} 85.0\% &\cellcolor[RGB]{174,255,174}  85.9\% &\cellcolor[RGB]{207,255,207}    83.5\% \\
        R+C+G      &\cellcolor[RGB]{226,255,226} 82.1\% &\cellcolor[RGB]{193,255,193} 84.5\% &\cellcolor[RGB]{221,255,221}  82.5\% &\cellcolor[RGB]{202,255,202} 83.9\% &\cellcolor[RGB]{178,255,178}  85.6\% &\cellcolor[RGB]{204,255,204}    83.7\% \\
        R+C+G+N    &\cellcolor[RGB]{202,255,202} 83.9\% &\cellcolor[RGB]{200,255,200} 84.0\% &\cellcolor[RGB]{230,255,230}  81.8\% &\cellcolor[RGB]{176,255,176} 85.8\% &\cellcolor[RGB]{155,255,155}  \textbf{87.3\%} &\cellcolor[RGB]{180,255,180}    85.5\% \\
        \hline
        \multicolumn{7}{|c|}{Tanker rail}\\
        \hline
        \backslashbox{clsr.}{detr.} &     R &   R+C &  R+C+G &   R+N &  R+C+N &  R+C+G+N \\
        \hline
        End-to-End &\cellcolor[RGB]{255,255,255} 39.7\% &\cellcolor[RGB]{255,255,255} 47.9\% &\cellcolor[RGB]{255,255,255}  46.7\% &\cellcolor[RGB]{255,255,255} 41.7\% &\cellcolor[RGB]{255,255,255}  46.2\% &\cellcolor[RGB]{255,255,255}    49.7\% \\
        R          &\cellcolor[RGB]{190,255,190} 54.3\% &\cellcolor[RGB]{197,255,197} 53.8\% &\cellcolor[RGB]{194,255,194}  54.0\% &\cellcolor[RGB]{196,255,196} 53.9\% &\cellcolor[RGB]{214,255,214}  52.7\% &\cellcolor[RGB]{214,255,214}    52.7\% \\
        R+N        &\cellcolor[RGB]{184,255,184} 54.7\% &\cellcolor[RGB]{205,255,205} 53.3\% &\cellcolor[RGB]{213,255,213}  52.8\% &\cellcolor[RGB]{194,255,194} 54.0\% &\cellcolor[RGB]{214,255,214}  52.7\% &\cellcolor[RGB]{214,255,214}    52.7\% \\
        R+G        &\cellcolor[RGB]{220,255,220} 52.3\% &\cellcolor[RGB]{235,255,235} 51.3\% &\cellcolor[RGB]{247,255,247}  50.5\% &\cellcolor[RGB]{219,255,219} 52.4\% &\cellcolor[RGB]{244,255,244}  50.7\% &\cellcolor[RGB]{238,255,238}    51.1\% \\
        R+G+N      &\cellcolor[RGB]{184,255,184} 54.7\% &\cellcolor[RGB]{207,255,207} 53.2\% &\cellcolor[RGB]{223,255,223}  52.1\% &\cellcolor[RGB]{190,255,190} 54.3\% &\cellcolor[RGB]{226,255,226}  51.9\% &\cellcolor[RGB]{228,255,228}    51.8\% \\
        R+C        &\cellcolor[RGB]{182,255,182} 54.8\% &\cellcolor[RGB]{205,255,205} 53.3\% &\cellcolor[RGB]{208,255,208}  53.1\% &\cellcolor[RGB]{188,255,188} 54.4\% &\cellcolor[RGB]{219,255,219}  52.4\% &\cellcolor[RGB]{213,255,213}    52.8\% \\
        R+C+N      &\cellcolor[RGB]{155,255,155} \textbf{56.6\%} &\cellcolor[RGB]{172,255,172} 55.5\% &\cellcolor[RGB]{176,255,176}  55.2\% &\cellcolor[RGB]{173,255,173} 55.4\% &\cellcolor[RGB]{187,255,187}  54.5\% &\cellcolor[RGB]{191,255,191}    54.2\% \\
        R+C+G      &\cellcolor[RGB]{196,255,196} 53.9\% &\cellcolor[RGB]{210,255,210} 53.0\% &\cellcolor[RGB]{200,255,200}  53.6\% &\cellcolor[RGB]{197,255,197} 53.8\% &\cellcolor[RGB]{229,255,229}  51.7\% &\cellcolor[RGB]{216,255,216}    52.6\% \\
        R+C+G+N    &\cellcolor[RGB]{188,255,188} 54.4\% &\cellcolor[RGB]{210,255,210} 53.0\% &\cellcolor[RGB]{211,255,211}  52.9\% &\cellcolor[RGB]{197,255,197} 53.8\% &\cellcolor[RGB]{214,255,214}  52.7\% &\cellcolor[RGB]{216,255,216}    52.6\% \\
        \hline
    \end{tabular}
    \label{b2n_lo_shot_results}
\end{table}

\begin{table}
    \centering
    \caption{AP scores broad-to-narrow vs end-to-end detection, zero-shot.}
    \tiny
    \begin{tabular}{|l|rrrr|rrrr|}
        \hline
		& \multicolumn{4}{c|}{Aircraft A}& \multicolumn{4}{c|}{Aircraft B}\\
        \hline
        \backslashbox{clsr.}{detr.} &     R &   R+C &   R+G &  R+C+G &     R &   R+C &   R+G &  R+C+G \\
        \hline
        End-to-End &     &\cellcolor[RGB]{255,255,255} 32.6\% &\cellcolor[RGB]{253,255,253} 70.4\% &\cellcolor[RGB]{233,255,233}  75.7\%  &     &\cellcolor[RGB]{255,255,255} 30.7\% &\cellcolor[RGB]{255,255,255} 34.5\% &\cellcolor[RGB]{255,255,255}  41.1\% \\
        C.plain    &\cellcolor[RGB]{255,255,255} 27.9\% &\cellcolor[RGB]{184,255,184} 88.0\% &\cellcolor[RGB]{198,255,198} 84.5\% &\cellcolor[RGB]{175,255,175}  90.5\%   &\cellcolor[RGB]{255,255,255} 60.0\% &\cellcolor[RGB]{193,255,193} 72.8\% &\cellcolor[RGB]{230,255,230} 65.2\% &\cellcolor[RGB]{228,255,228}  65.6\% \\
        C.both     &\cellcolor[RGB]{255,255,255} 33.7\% &\cellcolor[RGB]{173,255,173} 90.9\% &\cellcolor[RGB]{185,255,185} 87.8\% &\cellcolor[RGB]{175,255,175}  90.5\%   &\cellcolor[RGB]{227,255,227} 65.8\% &\cellcolor[RGB]{168,255,168} 78.0\% &\cellcolor[RGB]{207,255,207} 70.0\% &\cellcolor[RGB]{201,255,201}  71.1\% \\
        C.nnstx    &\cellcolor[RGB]{255,255,255} 32.8\% &\cellcolor[RGB]{170,255,170} 91.8\% &\cellcolor[RGB]{200,255,200} 83.9\% &\cellcolor[RGB]{174,255,174}  90.7\%   &\cellcolor[RGB]{207,255,207} 69.9\% &\cellcolor[RGB]{160,255,160} 79.6\% &\cellcolor[RGB]{194,255,194} 72.6\% &\cellcolor[RGB]{196,255,196}  72.2\% \\
        C+G.plain  &\cellcolor[RGB]{255,255,255} 32.9\% &\cellcolor[RGB]{183,255,183} 88.3\% &\cellcolor[RGB]{208,255,208} 81.9\% &\cellcolor[RGB]{180,255,180}  89.1\%   &\cellcolor[RGB]{222,255,222} 66.8\% &\cellcolor[RGB]{166,255,166} 78.4\% &\cellcolor[RGB]{212,255,212} 69.0\% &\cellcolor[RGB]{203,255,203}  70.7\% \\
        C+G.both   &\cellcolor[RGB]{255,255,255} 35.0\% &\cellcolor[RGB]{156,255,156} 95.3\% &\cellcolor[RGB]{180,255,180} 89.0\% &\cellcolor[RGB]{155,255,155}  \textbf{95.5\%}   &\cellcolor[RGB]{214,255,214} 68.5\% &\cellcolor[RGB]{162,255,162} 79.2\% &\cellcolor[RGB]{201,255,201} 71.1\% &\cellcolor[RGB]{193,255,193}  72.9\% \\
        C+G.nnstx  &\cellcolor[RGB]{255,255,255} 29.2\% &\cellcolor[RGB]{160,255,160} 94.2\% &\cellcolor[RGB]{193,255,193} 85.9\% &\cellcolor[RGB]{170,255,170}  91.8\%   &\cellcolor[RGB]{230,255,230} 65.1\% &\cellcolor[RGB]{171,255,171} 77.4\% &\cellcolor[RGB]{212,255,212} 69.0\% &\cellcolor[RGB]{200,255,200}  71.3\% \\
        G.plain    &\cellcolor[RGB]{255,255,255} 27.7\% &\cellcolor[RGB]{168,255,168} 92.2\% &\cellcolor[RGB]{202,255,202} 83.6\% &\cellcolor[RGB]{173,255,173}  90.8\%   &\cellcolor[RGB]{251,255,251} 60.8\% &\cellcolor[RGB]{175,255,175} 76.6\% &\cellcolor[RGB]{236,255,236} 63.9\% &\cellcolor[RGB]{215,255,215}  68.3\% \\
        G.both     &\cellcolor[RGB]{255,255,255} 29.1\% &\cellcolor[RGB]{160,255,160} 94.1\% &\cellcolor[RGB]{204,255,204} 83.0\% &\cellcolor[RGB]{167,255,167}  92.5\%   &\cellcolor[RGB]{217,255,217} 67.8\% &\cellcolor[RGB]{155,255,155} \textbf{80.7\%} &\cellcolor[RGB]{193,255,193} 72.8\% &\cellcolor[RGB]{191,255,191}  73.2\% \\
        G.nnstx    &\cellcolor[RGB]{255,255,255} 28.2\% &\cellcolor[RGB]{162,255,162} 93.7\% &\cellcolor[RGB]{201,255,201} 83.8\% &\cellcolor[RGB]{169,255,169}  92.0\%   &\cellcolor[RGB]{215,255,215} 68.2\% &\cellcolor[RGB]{157,255,157} 80.3\% &\cellcolor[RGB]{197,255,197} 72.1\% &\cellcolor[RGB]{191,255,191}  73.3\% \\
        \hline
		& \multicolumn{4}{c|}{Passenger rail}& \multicolumn{4}{c|}{Tanker rail}\\
        \hline
        \backslashbox{clsr.}{detr.} &     R &   R+C &   R+G &  R+C+G &     R &   R+C &   R+G &  R+C+G \\
        \hline
        End-to-End &     &\cellcolor[RGB]{255,255,255} 12.0\% &\cellcolor[RGB]{255,255,255}  6.4\% &\cellcolor[RGB]{255,255,255}  13.3\%  &     &\cellcolor[RGB]{255,255,255}  0.0\% &\cellcolor[RGB]{255,255,255}  0.0\% &\cellcolor[RGB]{255,255,255}   0.0\% \\
        C.plain    &\cellcolor[RGB]{255,255,255} 21.5\% &\cellcolor[RGB]{161,255,161} 44.9\% &\cellcolor[RGB]{189,255,189} 40.4\% &\cellcolor[RGB]{195,255,195}  39.5\%   &\cellcolor[RGB]{209,255,209}  9.9\% &\cellcolor[RGB]{164,255,164} 19.6\% &\cellcolor[RGB]{184,255,184} 15.4\% &\cellcolor[RGB]{167,255,167}  19.0\% \\
        C.both     &\cellcolor[RGB]{255,255,255} 21.4\% &\cellcolor[RGB]{164,255,164} 44.4\% &\cellcolor[RGB]{196,255,196} 39.4\% &\cellcolor[RGB]{201,255,201}  38.5\%   &\cellcolor[RGB]{205,255,205} 10.9\% &\cellcolor[RGB]{158,255,158} 21.0\% &\cellcolor[RGB]{184,255,184} 15.4\% &\cellcolor[RGB]{155,255,155}  \textbf{21.6\%} \\
        C.nnstx    &\cellcolor[RGB]{255,255,255} 19.2\% &\cellcolor[RGB]{182,255,182} 41.6\% &\cellcolor[RGB]{223,255,223} 35.0\% &\cellcolor[RGB]{222,255,222}  35.2\%   &\cellcolor[RGB]{204,255,204} 11.1\% &\cellcolor[RGB]{174,255,174} 17.4\% &\cellcolor[RGB]{186,255,186} 14.9\% &\cellcolor[RGB]{185,255,185}  15.2\% \\
        C+G.plain  &\cellcolor[RGB]{255,255,255} 20.0\% &\cellcolor[RGB]{155,255,155} \textbf{45.8\%} &\cellcolor[RGB]{194,255,194} 39.7\% &\cellcolor[RGB]{206,255,206}  37.7\% &\cellcolor[RGB]{217,255,217}  8.2\% &\cellcolor[RGB]{202,255,202} 11.4\% &\cellcolor[RGB]{206,255,206} 10.5\% &\cellcolor[RGB]{199,255,199}  12.2\% \\
        C+G.both   &\cellcolor[RGB]{255,255,255} 17.5\% &\cellcolor[RGB]{200,255,200} 38.7\% &\cellcolor[RGB]{233,255,233} 33.4\% &\cellcolor[RGB]{239,255,239}  32.6\%  &\cellcolor[RGB]{231,255,231}  5.1\% &\cellcolor[RGB]{215,255,215}  8.6\% &\cellcolor[RGB]{220,255,220}  7.6\% &\cellcolor[RGB]{212,255,212}   9.3\% \\
        C+G.nnstx  &\cellcolor[RGB]{255,255,255} 18.7\% &\cellcolor[RGB]{197,255,197} 39.2\% &\cellcolor[RGB]{233,255,233} 33.4\% &\cellcolor[RGB]{242,255,242}  32.0\%  &\cellcolor[RGB]{240,255,240}  3.2\% &\cellcolor[RGB]{237,255,237}  3.8\% &\cellcolor[RGB]{241,255,241}  3.1\% &\cellcolor[RGB]{235,255,235}   4.3\% \\
        G.plain    &\cellcolor[RGB]{255,255,255} 15.6\% &\cellcolor[RGB]{212,255,212} 36.8\% &\cellcolor[RGB]{254,255,254} 30.1\% &\cellcolor[RGB]{255,255,255}  29.5\%  &\cellcolor[RGB]{243,255,243}  2.5\% &\cellcolor[RGB]{241,255,241}  3.0\% &\cellcolor[RGB]{243,255,243}  2.6\% &\cellcolor[RGB]{238,255,238}   3.6\% \\
        G.both     &\cellcolor[RGB]{255,255,255} 18.2\% &\cellcolor[RGB]{193,255,193} 39.8\% &\cellcolor[RGB]{232,255,232} 33.6\% &\cellcolor[RGB]{240,255,240}  32.4\%  &\cellcolor[RGB]{250,255,250}  1.0\% &\cellcolor[RGB]{247,255,247}  1.8\% &\cellcolor[RGB]{248,255,248}  1.5\% &\cellcolor[RGB]{244,255,244}   2.3\% \\
        G.nnstx    &\cellcolor[RGB]{255,255,255} 13.4\% &\cellcolor[RGB]{231,255,231} 33.8\% &\cellcolor[RGB]{255,255,255} 27.2\% &\cellcolor[RGB]{255,255,255}  27.0\%  &\cellcolor[RGB]{246,255,246}  1.9\% &\cellcolor[RGB]{244,255,244}  2.4\% &\cellcolor[RGB]{245,255,245}  2.2\% &\cellcolor[RGB]{240,255,240}   3.2\% \\
        \hline

    \end{tabular}
    \label{b2n_zero_shot_results}
\end{table}

\subsection{Analysis of Broad-to-Narrow}

\begin{table}
    \centering
    \caption{Maximum detection model recall with imprecision 100 or better. Broad detection models vs. end-to-end models.}
    \tiny
    \begin{tabular}{|l|rrrr|}
        \hline
        \backslashbox{detr.}{class} &  Aircraft A   &  Aircraft B  &  Passenger & Tanker \\
        \hline
        \multicolumn{5}{|c|}{Zero-shot}\\
        \hline
        Br. R &  52.4\% &    94.4\% &      71.0\% &   43.1\% \\
        Br. R+C &  97.6\% &    98.9\% &      86.6\% &   56.9\% \\
        Br. R+G & 100.0\% &    96.7\% &      86.6\% &   59.6\% \\
        Br. R+C+G & 100.0\% &    98.3\% &      86.2\% &   58.0\% \\
        \hline
        E2E C &  61.9\% &    52.2\% &      32.3\% &    0.0\% \\
        E2E G &  88.1\% &    70.0\% &      46.9\% &    0.0\% \\
        E2E C+G &  85.7\% &    69.4\% &      44.3\% &    0.5\% \\ 
        \hline
        \multicolumn{5}{|c|}{Low-shot}\\
        \hline
        B2N R  &  76.2\% &    97.8\% &      96.3\% &   63.3\% \\
        B2N R+C  &  95.2\% &   100.0\% &      96.3\% &   65.4\% \\
        B2N R+C+G  & 100.0\% &    99.4\% &      93.7\% &   65.4\% \\
        B2N R+N &  92.9\% &    96.7\% &      97.1\% &   66.0\% \\
        B2N R+C+N &  97.6\% &    99.4\% &      97.0\% &   62.8\% \\
        B2N R+C+G+N & 100.0\% &    99.4\% &      95.1\% &   61.7\% \\
        \hline
        E2E R  &  90.5\% &    97.8\% &      95.2\% &   55.9\% \\
        E2E R+C  &  83.3\% &    98.9\% &      94.3\% &   64.4\% \\
        E2E R+C+G  & 100.0\% &    98.9\% &      92.2\% &   63.3\% \\
        E2E R+N &  88.1\% &    95.6\% &      96.3\% &   48.9\% \\
        E2E R+C+N &  92.9\% &    98.3\% &      95.6\% &   58.5\% \\
		E2E R+C+G+N & 100.0\% &    98.9\% &      93.2\% &   65.4\% \\   
        \hline
    \end{tabular}
    
    \label{maxrec}
\end{table}

One of the shortcomings of the end-to-end approach to narrow class, zero-shot detection problems is that models trained with low or zero samples of the target class often have disappointing levels of recall, even with the minimum confidence threshold. In the \textit{broad-to-narrow} meta-architecture, it falls to the broad stage to fix this problem. By training the broad stage with more positive, real examples from sibling classes, we reason that the detector will have a higher level of absolute recall. 

To check this hypothesis and provide a better understanding of the nature of \textit{broad-to-narrow's} improved performance, we look at the recall levels of the broad detector's components compared to the recall levels of the end-to-end models in the zero-shot setting. 
We could examine the absolute recall (at the minimum score threshold made available by Faster R-CNN), but some of the rail car models become too imprecise at the very low threshold realm. 
Instead we compute the maximum recall given the model is producing no more than one hundred false detects per true detect (precision = $1/101$).
Table \ref{maxrec} shows these values and demonstrates that, to a large degree in the zero-shot setting but to a lesser degree in the low-shot setting, the broad stage detectors have high recall levels compared to their end-to-end counterparts.

\section{Conclusion and Future Work}\
\label{conclusion}

In this work we implemented a comprehensive synthetic data pipeline including extensive testing of neural style transfer and our own GAN reskinning system. These data sets were used to train object detection models on challenging geospatial analysis problems as well as to train our powerful \textit{broad-to-narrow} meta-model.
Our experiments show that our hierarchical strategy is valuable and even essential in the zero-shot problems.
The effect of our various synthetic data sources seems to be quite class dependent, but the GAN-Reskinner provides a consistent benefit in training traditional detectors end-to-end.
Our hierarchical detection strategy allows us to do separate training data optimization for the detection and classification tasks.

The \textit{broad-to-narrow} framework shows promise, and as a future work we aim to expand on that modular technology. We would like to investigate the effect of adding an auxiliary false detect filter to decrease the number of background false detects.
We also observe the \textit{broad-to-narrow} meta-model as a strategy to easily integrate recent advances in image classification (adversarial training or ensembling, for example) to our detection problems.

Recall the notation $X$, $R(X)$, $X'$ for image, edge information, and synthetic image in our reskinning system.
The system relies on 
\begin{enumerate}
    \item the GAN generator $G$'s being able to learn a reasonable distributional approximation of $X$ given $R(X)$
    \item the representation function $R$'s ability to reduce any artifacts in a synthetic image $X'$.
\end{enumerate}
The chosen representation functions ignore color information, correspondingly, $G$ can struggle to reproduce vivid, lifelike color. We would like to explore modifications to $R$ which pass some information about color but preserve the second property above.

Rather than relying on the GAN mapping $X'\mapsto G(R(X'))$ to map onto a subspace of realistic images, we can explore direct targeting of the synthetic/real gap with image level adversarial domain adaptation (SimGAN \cite{simgan}). 
We hypothesize that the SimGAN refinement strategy coupled with some form of localization can improve synthetic overhead training images.

\bibliographystyle{ACM-Reference-Format}
\bibliography{synth_b2n}

\end{document}